\documentclass{article}
\usepackage{ijcai20}
\usepackage{times}

\usepackage{soul}
\usepackage{url}
\usepackage[hidelinks]{hyperref}
\usepackage[utf8]{inputenc}
\usepackage[small]{caption}
\usepackage{graphicx}
\usepackage{amsmath}
\usepackage{amssymb}
\usepackage{booktabs}
\usepackage{tikz}
\usetikzlibrary{calc,intersections,arrows,arrows.meta,3d,decorations.text,shapes.arrows,fit,backgrounds,shapes,snakes}

\newcommand{\tikzmark}[1]{\tikz[overlay,remember picture] \node (#1) {};}

\title{Graph Neural Networks Meet Neural-Symbolic Computing:\\ A Survey and Perspective}

\author{
Luis C. Lamb$^1$\and 
Artur d'Avila Garcez$^2$\and 
Marco Gori$^{3,4}$\and 
Marcelo O.R. Prates$^1$\and\\
Pedro H.C. Avelar$^{1,3}$\and
Moshe Y. Vardi$^5$
\affiliations
$^1$UFRGS, Brazil 
$^2$City, University of London
$^3$University of Siena, IT \\
$^4$Université Côte d'Azur, 3IA, FR
$^5$Rice University, Houston, USA
\emails
\{lamb, morprates, phcavelar\}@inf.ufrgs.br,
a.garcez@city.ac.uk,
marco.gori@unisi.it,
vardi@cs.rice.edu
}

\begin{document}

\maketitle

\begin{abstract}
Neural-symbolic computing has now become the subject of interest of both  academic and industry research laboratories. 
Graph Neural Networks (GNNs) have been widely used in relational and symbolic domains, with widespread application of GNNs in combinatorial optimization, constraint satisfaction, relational reasoning and other scientific domains.
The need for improved explainability, interpretability and trust of AI systems in general demands principled methodologies, as suggested by neural-symbolic computing. 
In this paper, we review the state-of-the-art on the use of GNNs as a model of neural-symbolic computing. This includes the application of GNNs in several  domains as well as their relationship to current developments in neural-symbolic computing. 
\end{abstract}
\section{Introduction}
Over the last decade Artificial Intelligence in general, and deep learning in particular, have been the focus of intensive research endeavors, gathered media attention and led to debates on their impacts both in academia and industry \cite{marcus2020,raghavan19}. The recent AI Debate in Montreal with Yoshua Bengio and Gary Marcus \cite{marcus2020}, and the AAAI-2020 fireside conversation with Nobel Laureate Daniel Kahneman and the 2018 Turing Award winners and deep learning pioneers Geoff Hinton, Yoshua Bengio and Yann LeCun have led to new perspectives on the future of AI. 
It has now been  argued 
that if one aims to build richer AI systems, i.e. semantically sound, explainable, and reliable, one has to add a sound reasoning layer to deep learning \cite{marcus2020}. Kahneman has made this point clear when he stated at AAAI-2020 that \emph{``...so far as I'm concerned, System 1 certainly knows language... System 2... \textbf{does involve certain manipulation of symbols}."} \cite{fireside2020}. 

Kahneman's comments address recent parallels made by AI researchers between ``Thinking, Fast and Slow" and the so-called ``AI's systems 1 and 2", which could, in principle, be modelled by deep learning and symbolic reasoning, respectively.\footnote{``Thinking, Fast and Slow", by Daniel Kahneman: New York, FSG, 2011, describes the author's \emph{``... current understanding of judgment and decision making, which has been shaped by psychological discoveries of recent decades."}}   
In this paper, we present a survey and relate recent research results on:
(1) Neural-Symbolic Computing, by summarizing the main approaches to rich knowledge representation and reasoning within deep learning, and 
(2) the approach pioneered by the authors and others of Graph Neural Networks (GNN) for learning and reasoning about problems that require relational structures or symbolic learning. 
Although recent papers have surveyed  GNNs, including  \cite{battaglia2018relational,chami2020,wu2019comprehensive,Zhang2018} they have not focused on the relationship between GNNs and neural-symbolic computing (NSC). \cite{bengio2018tourdhorizon} also touches particular topics related to some we discuss here, in particular to do with meta-transfer learning.
Recent surveys in neural-symbolic computing \cite{garcez2015,JAL19,Joe19} have not exploited the highly relevant applications of GNN in symbolic and relational learning, or the relationship between the two approaches.

\paragraph{Our Contribution}As mentioned above, recent work have surveyed graph neural networks and neural-symbolic computing, but to the best of our knowledge, no survey has reviewed and analysed the recent results on the specific relationship between GNNs and NSC. We also outline the promising directions for research and applications combining GNNs and NSC from the perspective of symbolic reasoning tasks. The above-referenced surveys on GNNs, although comprehensive, all describe other application domains. 
The remainder of the paper is organized as follows.
In Section 2, we present an overview and taxonomy of neural-symbolic computing. In Section 3, we discuss the main GNN models and their relationship to neural-symbolic computing. We then outline the main GNN architectures and their use in relational and symbolic learning. Finally, we conclude and point out directions for further research. We shall assume familiarity with neural learning and symbolic AI.

\section{Neural-Symbolic Computing Taxonomy}
At this year's Robert S. Engelmore Memorial Lecture, at the AAAI Conference on Artificial Intelligence, New York, February 10th, 2020, Henry Kautz introduced a taxonomy for neural-symbolic computing as part of a talk entitled \emph{The Third AI Summer}. Six types of neural-symbolic systems are outlined: 1. \textsc{symbolic Neuro symbolic}, 2. \textsc{Symbolic[Neuro]}, 3. \textsc{Neuro;Symbolic}, 4. \textsc{Neuro:Symbolic $\rightarrow$ Neuro}, 5. \textsc{Neuro\textsubscript{Symbolic}} and 6. \textsc{Neuro[Symbolic]}. 

The origin of GNNs \cite{scarselli2008graph} can be traced back to neural-symbolic computing (NSC) in that both sought to enrich the vector representations in the inputs of neural networks, first by accepting tree structures and then graphs more generally. In this sense, according to Kautz's taxonomy, GNNs are a \textsc{type 1} neural-symbolic system. 
GNNs \cite{battaglia2018relational} were recently combined with convolutional networks in novel ways which have produced impressive results on data efficiency. In parallel, NSC has focused on the learning of adequate embeddings for the purpose of symbolic computation. 
This branch of neural-symbolic computing, which includes Logic Tensor Networks \cite{LTN} and Tensor Product Representations \cite{Smolensky} has been called in \cite{JAL19} \emph{tensorization} methods and draw similarities with \cite{DiligentiGS17} that use fuzzy methods in representing first-order logic. These have been classified by Kautz as \textsc{type 5} neural-symbolic systems, as also discussed in what follows. A natural point of contact between GNNs and NSC is the provision of rich embeddings and attention mechanisms towards structured reasoning and efficient learning. 

\textsc{Type 1} neural-symbolic integration is standard deep learning, which some may argue is a stretch to refer to as neural-symbolic, but which is included here to note that the input and output of a neural network can be made of symbols e.g. in the case of language translation or question answering applications. \textsc{Type 2} are hybrid systems such as DeepMind's AlphaGo and other systems, where the core neural network is loosely-coupled with a symbolic problem solver such as Monte Carlo tree search. \textsc{Type 3} is also a hybrid system whereby a neural network focusing on one task (e.g. object detection) interacts via input/output with a symbolic system specialising in a complementary task (e.g. query answering). Examples include the neuro-symbolic concept learner \cite{Mao_2019} and deepProbLog \cite{Robin_2018,kersting2019}. 

In a \textsc{type 4} neural-symbolic system, symbolic knowledge is compiled into the training set of a neural network. Kautz offers \cite{Lample2020Deep} as an example. Here, we would also include other tightly-coupled neural-symbolic systems where various forms of symbolic knowledge, not restricted to \emph{if-then} rules only, can be translated into the initial architecture and set of weights of a neural network \cite{garcez_book2}, in some cases with guarantees of correctness. We should also mention \cite{ArabshahiSA18}, which learn and reason over mathematical constructions, as well as \cite{Arabshahi19}, which propose a learning architecture that extrapolates to much harder symbolic maths reasoning problems than what was seen during training. \textsc{Type 5} are those tightly-coupled neural-symbolic systems where a symbolic logic rule is mapped onto a distributed representation (an embedding) and acts as a soft-constraint (a regularizer) on the network's loss function. Examples of these are \cite{Smolensky} and \cite{LTN}.

Finally, \textsc{type 6} systems should be capable, according to Kautz, of \emph{true symbolic reasoning inside a neural engine}. It is what one could refer to as a fully-integrated system. Early work in neural-symbolic computing has achieved this: see \cite{garcez_book2} for a historical overview; and some \textsc{type 4} systems are also capable of it \cite{garcez_book2,garcez2015,Hitzler04}, but in a localist rather than a distributed architecture and using simpler forms of embedding than \textsc{type 5} systems. Kautz adds that \textsc{type 6} systems should be capable of \emph{combinatorial reasoning}, suggesting using an attention schema to achieve it effectively. In fact, attention mechanisms can be used to solve graph problems, for example with pointer networks \cite{vinyals2015pointer}. It should be noted that the same problems can be solved through other NSC architectures, such as GNNs \cite{prates2019tsp}. This idea resonates with the recent proposal outlined by Bengio in the AI debate of December 2019. 
  
In what concerns neural-symbolic computing theory, the study of \textsc{type 6} systems is highly relevant. In practical terms, a tension exists between effective learning and sound reasoning, which may prescribe the use of a more hybrid approach (\textsc{types} 3 to 5) or variations thereof such as the use of attention with tensorization. Orthogonal to the above taxonomy, but mostly associated so far with \textsc{type 4}, is the study of the limits of reasoning within neural networks w.r.t. full first-order, higher-order and non-classical logic theorem proving \cite{NIPS03,garcez2015}. In this paper, as we revisit the use of rich logic embeddings in \textsc{type 5} systems, notably Logic Tensor Networks \cite{LTN}, alongside the use of attention mechanisms or convolutions in GNNs, we will seek to propose a research agenda and specific applications of symbolic reasoning and statistical learning towards the sound development of \textsc{type 6} systems.

\section{Graph Neural Networks Meet Neural-Symbolic Computing}
One of the key concepts in machine learning is that of \textbf{priors} or \textbf{inductive biases} -- the set of assumptions that a learner uses to compute predictions on test data. In the context of deep learning (DL), the design of neural building blocks that enforce strong priors has been a major source of breakthroughs. For instance, the priors obtained through feedforward layers encourage the learner to combine features additively, while the ones obtained through dropout discourage it to overfit and the ones obtained through multi-task learning encourage it to prefer sets of parameters that explain more than one task. 

One of the most influential neural building blocks, having helped pave the way for the DL revolution, is the convolutional layer \cite{Hinton-nature}. Convolutional architectures are successful for tasks defined over Euclidean signals because they enforce equivariance to spatial translation. This is a useful property to have when learning representations for objects regardless of their position in a scene. 

Analogously, recurrent layers enforce equivariance in time that is useful for learning over sequential data. Recently, attention mechanisms, through the advent of transformer networks, have enabled advancing the state-of-art in many sequential tasks, notably in natural language processing \cite{devlin2018bert,goyal2019} and symbolic reasoning tasks such as solving math equations and integrals\footnote{It is advisable to read \cite{Lample2020Deep} alongside this critique of its limitations \cite{ernie}} \cite{Lample2020Deep}. Attention encourages the learner to combine representations additively, while also enforcing permutation invariance. All three architectures take advantage of \textbf{sparse connectivity} -- another important design in DL which is key to enable the training of larger models. Sparse connectivity and neural, building blocks with strong priors usually go hand in hand, as the latter leverage symmetries in the input space to cut down parameters through invariance to different types of transformations.
\begin{figure}[t]
\begin{center}
    $f(\mathbf{x}) = (x_1 \vee {\neg \tikzmark{a} x_5} \vee x_2 \vee \tikzmark{b}{x_3} \vee \neg x_4)$
  \begin{tikzpicture}[overlay,remember picture, distance=0.16cm,scale=2]
    \draw[very thick,<->,>=stealth,black, bend right=-30] ($(a.center) - (0,-0.2)$) to ($(b.center) - (0,-0.2)$);
  \end{tikzpicture}
\caption{Due to permutation invariance, literals $\neg x_5$ and $x_3$ can exchange places with no effect to the boolean function $f(\mathbf{x})$. There are $5! = 120$ such permutations.}
\label{fig:CNF-exchange}
\end{center}
\end{figure}
NSC architectures often combine the key design concepts from convolutional networks and attention-based architectures to enforce \textbf{permutation invariance} over the elements of a set or the nodes of a graph (Fig. \ref{fig:CNF-exchange}). Some NSC architectures such as Pointer Networks \cite{vinyals2015pointer} implement attention directly over a set of inputs $\mathbf{X} = \{x_1, \dots, x_n\}$ coupled with a decoder that outputs a sequence $(i_1, i_2, \dots i_n) \in [1,n]^{m}$ of ``pointers'' to the input elements (hence the name). Note that both formalizations are defined over \textbf{set} inputs rather than \textbf{sequential} ones.
\subsection{Logic Tensor Networks}
Tensorisation is a class of approaches that embeds first-order logic (FOL) symbols such as constants, facts and rules into real-valued tensors. Normally, constants are represented as one-hot vectors (first-order tensor). Predicates and functions are matrices (second-order tensor) or higher-order tensors.

In early work, embedding techniques were
proposed to transform symbolic representations into vector spaces where
reasoning can be done through matrix computation
\cite{Bordes_2011,LTN,Santoro_2017}. Training
embedding systems can be carried out as distance learning using backpropagation. Most research in this direction focuses on
representing relational predicates in a neural network. This is known
as ``relational embedding"
\cite{Ilya_2008}. For representation
of more complex logical structures, i.e. FOL formulas, a system named {\it Logic Tensor Network} (LTN) \cite{LTN} is proposed by extending {\it Neural Tensor Networks} (NTN), a state-of-the-art relational embedding method. LTNs effectively implement learning using symbolic information as a prior, as pointed out by \cite{van2019boxology}. Related ideas are discussed formally in the context of constraint-based learning and reasoning \cite{JAL19}. Recent research in first-order logic programs has successfully exploited advantages of distributed representations of logic symbols for efficient reasoning, inductive programming  \cite{Evans_18} and differentiable theorem proving \cite{Rocktaschel_2016}.

\subsection{Pointer Networks}
The Pointer Network (PN) formalization \cite{vinyals2015pointer} is a neural architecture meant for computing a $m$-sized \textbf{sequence} $(i_1, i_2, \dots i_n) \in [1, n]^m$ over the elements of an input \textbf{set} $\mathbf{X} = \{x_1, \dots, x_n\}$. PN implements a simple modification over the traditional \emph{seq2seq} model, augmenting it with a simplified variant of the attention mechanism whose outputs are interpreted as ``pointers'' to the input elements. Traditional \emph{seq2seq} models implement an encoder-decoder architecture in which the elements of the input sequence are consumed in order and used to update the encoder's hidden state at each step. Finally, a decoder consumes the encoder's hidden state and is used to yield a sequence of outputs, one at a time. 

It is known that \emph{seq2seq} models tend to exhibit improved performance when augmented with an attention mechanism, a phenomenon noticeable from the perspective of Natural Language Processing  \cite{devlin2018bert}. Traditional models however yield sequences of outputs over a \textbf{fixed-length dictionary} (for instance a dictionary of tokens for language models), which is not useful for tasks whose output is defined over the input set and hence require a \textbf{variable-length dictionary}.
PN tackle this problem by encoding the $n$-sized input set $\mathcal{P}$ with a traditional encoding architecture and decoding a probability distribution $p(C_i | C_1, \dots C_{i-1}, \mathcal{P})$ over the set $\{1, \dots, n\}$ of indices at each step $i$ by computing a softmax over an attention layer parameterized by matrices $\mathbf{W_1}, \mathbf{W_2}$ and vector $\mathbf{v}$ feeding on the decoder state $d_i$ and the encoder states $e_i ~~ (1, \dots, n)$:
\begin{equation}
\begin{aligned}
%\begin{center}
    u_j^i = \mathbf{v}^\intercal \tanh \left(\mathbf{W_1}e_j \mathbf{W_2}d_i \right) ~~ j \in (1, \dots n) \\
   p(C_i | C_1, \dots C_{i-1}, \mathcal{P}) = \textrm{softmax}(u^i)
%\end{center} 
\end{aligned}
\end{equation}

The output pointers can then be used to compute loss functions over combinatorial optimization problems. In the original paper the authors define a PN to solve the Traveling Salesperson Problem (TSP) in which a beam search procedure is used to select cities given the probability distributions computed at each step and finally a loss function can computed for the output tour by adding the corresponding city distances.
Given their discrete nature, PNs are naturally suitable for many combinatorial problems (the original paper evaluates PN on Delauney Triangulation, TSP and Convex Hull problems). Unfortunately, even though PNs can solve problems over sets, they cannot be directly applied to general (non-complete) graphs.

\subsection{Convolutions as Self-Attention}
\label{GNN-architecture}
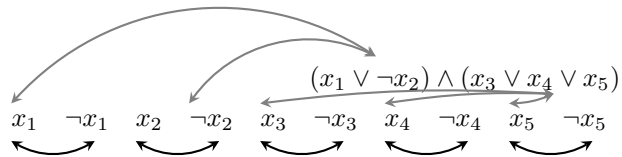
\begin{figure}[t]
\begin{center}
$ \begin{aligned}
    \\\\
    (x_1 \vee \tikzmark{c1} \neg x_2) \wedge (x_3 \vee x_4 \tikzmark{c2} \vee x_5) \\
    \begin{array}{cccccccccc} \tikzmark{x1} x_1 & \tikzmark{nx1} \neg x_1 & \tikzmark{x2} x_2 & \tikzmark{nx2} \neg x_2 & \tikzmark{x3} x_3 & \tikzmark{nx3} \neg x_3 & \tikzmark{x4} x_4 & \tikzmark{nx4} \neg x_4 & \tikzmark{x5} x_5 & \tikzmark{nx5} \neg x_5 \end{array}
    %\begin{array}{cccccccccc} \tikzmark{x1} x_1 & \tikzmark{x2} x_2 & \tikzmark{x3} x_3 & \tikzmark{x4} x_4 & \tikzmark{x5} x_5 & \neg \tikzmark{nx1} x_1 & \neg \tikzmark{nx2} x_2 & \neg \tikzmark{nx3} x_3 & \neg \tikzmark{nx4} x_4 & \neg \tikzmark{nx5} x_5 \end{array}
  \begin{tikzpicture}[overlay,remember picture,scale=2]
    \draw[thick,<->,>=stealth,gray, bend right = 40] ($(c1.center) + (0,0.2)$) to ($(x1.center) + (0,0.15)$);
    \draw[thick,<->,>=stealth,gray, bend right = 40] ($(c1.center) + (0,0.2)$) to ($(nx2.center) + (0,0.15)$);
    \draw[thick,<->,>=stealth,gray, bend right = 5] ($(c2.center) - (0,0.05)$) to ($(x3.center) + (0,0.15)$);
    \draw[thick,<->,>=stealth,gray, bend right = 5] ($(c2.center) - (0,0.05)$) to ($(x4.center) + (0,0.15)$);
    \draw[thick,<->,>=stealth,gray, bend left = 10] ($(c2.center) - (0,0.05)$) to ($(x5.center) + (0,0.15)$);
    \draw[thick,<->,>=stealth,black, bend right = 30] ($(x1.center) - (0,0.1)$) to ($(nx1.center) + (0.2,-0.1)$);
    \draw[thick,<->,>=stealth,black, bend right = 30] ($(x2.center) - (0,0.1)$) to ($(nx2.center) + (0.2,-0.1)$);
    \draw[thick,<->,>=stealth,black, bend right = 30] ($(x3.center) - (0,0.1)$) to ($(nx3.center) + (0.2,-0.1)$);
    \draw[thick,<->,>=stealth,black, bend right = 30] ($(x4.center) - (0,0.1)$) to ($(nx4.center) + (0.2,-0.1)$);
    \draw[thick,<->,>=stealth,black, bend right = 30] ($(x5.center) - (0,0.1)$) to ($(nx5.center) + (0.2,-0.1)$);
  \end{tikzpicture}
\end{aligned}$
\end{center}
\caption{CNF formula $F = (x_1 \vee \neg x_2) \wedge (x_3 \vee x_4 \vee x_5)$ represented as a graph: clauses and literals correspond to nodes, edges between clauses and literals are painted {\color{gray}gray} and edges between literals and their complements are painted {\color{black}black}}
\label{fig:CNF-as-graph}
\end{figure}

The core building block of models in the GNN family is the \textbf{graph convolution} operation, which is a neural building block that enables one to perform learning over graph inputs. Empowering DL architectures with the capacity of feeding on graph-based data is particularly suitable for neural-symbolic reasoning, as symbolic expressions are easily represented as graphs (Fig. \ref{fig:CNF-as-graph}). Furthermore, graph representations have useful properties such as permutation invariance and flexibility for generalization over the input size (models in the graph neural network family can be fed with graphs regardless of their size in terms of number of vertices). Graph convolutions can be seen as a variation of the well-known attention mechanism \cite{garcia18}. A graph convolution is essentially an attention layer with two key differences:
\begin{enumerate}
\item \textbf{There is no dot-product} for computing weights: encodings are simply added together with unit weights.\footnote{The Graph Attention network (GAT) however generalizes graph convolutions with dot-product attention \cite{velivckovic2017graph}}
\item The sum is masked with an adjacency mask, or in other words the graph convolution \textbf{generalizes attention for non-complete graphs}.
\end{enumerate}

All models in the GNN family learn continuous representations for graphs by embedding nodes into hyper-dimensional spaces, an insight motivated by graph embedding algorithms. A graph embedding corresponds to a function
%\begin{equation}
  $  f : \mathbf{V} \to \mathbb{R}^n  $ 
%\end{equation}
mapping from the set of vertices $\mathbf{V}$ of a graph $\mathbf{G} = (\mathbf{V}, \mathbf{E})$ to n-dimensional vectors.
In the context of GNNs, we are interested in learning the parameters $\theta$ of a function 
%\begin{equation}
  $  f: \mathcal{G} \times \theta \to (\mathbf{V} \to \mathbb{R}^n)$. 
%\end{equation}
That is, a parameterized function $f(\mathcal{G}, \theta)$ over the set of graphs $\mathcal{G}$ whose outputs are mappings $\mathbf{V} \to \mathbb{R}^n$ from vertices to n-dimensional vectors. In other words, \textbf{GNNs learn functions to encode vertices} in a generalized way. Note that since the output from a GNN is itself a function, there are no limitations for the number of vertices in the input graph. This useful property stems from the modular architecture of GNNs, which will be discussed at length in the sequel. We argue that this should be interesting to explore in the context of neural-symbolic computing in the representation and manipulation of variables within neural networks.

Generally, instead of synthesizing a vertex embedding function from the ground up, GNNs choose an initial, simpler vertex embedding such as mapping each vertex to the same (learned) vector representation or sampling vectors from a multivariate normal distribution, and then learn to \textbf{refine} this representation by iteratively updating representations for all vertices. The refinement process, which consists of each vertex aggregating information from its direct neighbors to update its own embedding is at the core of how GNNs learn properties over graphs. Over many refinement steps, vertices can aggregate structural information about progressively larger reachable subsets of the input graph. However we rely on a well-suited transformation at each step to enable vertices to make use of this structural information to solve problems over graphs. The graph convolution layer, described next in Section \ref{GCN}, implements such transformation.
\subsection{Graph Convolutional Networks}
\label{GCN}
Graph convolutions are defined in analogy to convolutional layers over Euclidean data. Both architectures compute weighted sums over a neighborhood. For CNNs, this neighborhood is the well known 9-connected or 25-connected neighborhood defined over pixels. One can think of the set of pixels of an image as a graph with a grid topology in which each vertex is associated with a vector representation corresponding to the Red/Green/Blue channels. The internal activations of a CNN can also be thought of graphs with grid topologies, but the vector representations for each pixel are generally embedded in spaces of higher dimensionality (corresponding to the number of convolutional kernels learned at each layer).
In this context, Graph Convolutional Networks (GCNs) \cite{kipf2016semi} can be thought of as a generalization of CNNs for non-grid topologies. Generalizing CNNs this way is tricky because one cannot rely anymore on learning $3 \times 3$ or $5 \times 5$ kernels, for two reasons:
\begin{enumerate}
\item In grid topologies pixels are embedded in 2-dimensional Euclidean space, which enables one to learn a specific weight for each neighbor on the basis of its relative position (left, right, central, top-right, etc.). This is not true for general graphs, and hence weights such as $\mathbf{W_{1,0}}, \mathbf{W_{1,1}}, \mathbf{W_{1,1}}, \mathbf{W_{0,1}}$ do not always have a clear interpretation.
\item In grid topologies each vertex has a fixed number of neighbors and weight sharing, but there is no such constraint for general graphs. Thus we cannot hope to learn a specific weight for each neighbor as the required number of such weights will vary with the input graph.
\end{enumerate} 
GCNs tackle this problem the following way: Instead of learning kernels corresponding to matrices of weights, they learn transformations for vector representations (embeddings) of graph vertices. Concretely, given a graph ${\mathbf{G} = (\mathbf{V}, \mathbf{E})}$ and a matrix ${\mathbf{x}^{(k)} \in \mathbb{R}^{|\mathbf{V}| \times d}}$ of vertex representations (i.e. $\vec{\mathbf{x_i}}^{(k)}$ is the vector representation of vertex $i$ at the k-th layer), a GCN computes the representations $\vec{\mathbf{x_i}}^{(k+1)}$ of vertex $i$ in the next layer as:
\begin{equation}\label{eq:GCN-1}
    \vec{\mathbf{x_i}}^{(k+1)} = \sigma\biggl
    (\sum_{j \in \mathcal{N}(i) \cup \{i\}}{\frac{\mathbf{\theta_k} \cdot \vec{\mathbf{x_j}}^{(k)}}{\sqrt{\deg(i)}\sqrt{\deg(j)}}}
    \biggr)
\end{equation}
In other words, we linearly transform the vector representation of each neighbor $j$ by multiplying it with a learned matrix of weights $\mathbf{\theta_k}$, normalizing it by the square roots of the degrees $\deg i, \deg j$ of both vertices, aggregate all results additively and finally apply a non-linearity $\sigma$. Note that $\mathbf{\theta_k}$ denotes the learned weight matrix for GCN layer $k$: in general one will stack $n$ different GCN layers together and hence learn the parameters of $n$ such matrices. Also note that one iterates over an extended neighborhood $\mathcal{N}(i) \cup \{i\}$, which includes $i$ itself. This is done to prevent ``forgetting'' the representation of the vertex being updated.
Equation \ref{eq:GCN-1} can be summarized as: 
%\begin{equation}\label{eq:GCN-2}
    $\mathbf{x}^{(k+1)} = \bigl(\mathbf{\tilde{D}}^{-\frac{1}{2}} \mathbf{\tilde{A}} \mathbf{\tilde{D}}^{-\frac{1}{2}} \bigr) \mathbf{x}^{(k)} \mathbf{\theta}^{(k)}$, 
%\end{equation}
where ${\mathbf{\tilde{A}} = \mathbf{A} + \mathbf{I}}$ is the adjacency matrix $\mathbf{A}$ plus self-loops ($\textbf{I}$ is the identity matrix) and ${\mathbf{\tilde{D}}}$ is the degree matrix of ${\mathbf{\tilde{A}}}$.
\subsection{The Graph Neural Network Model}
Although GCNs are conceptually simpler, the GNN model predates them by a decade, having been originally proposed by \cite{scarselli2008graph}. The model is similar to GCNs, with two key differences:\\
%\begin{enumerate} 
\textbf{(1)} One does not stack multiple independent layers as in GCNs. A single parameterized function is iterated many times, in analogy to recurrent  networks, until convergence.\\
\textbf{(2)} The transformations applied to neighbor vertex representations are not necessarily linear, and can be implemented by deep neural networks (e.g. by a multilayer perceptron).
%\end{enumerate}

Concretely, the graph neural network model defines parameterized functions $h: \mathbb{N} \times \mathbb{N} \times \mathbb{N} \times \mathbb{R}^d \to \mathbb{R}^d$ and $g: \mathbb{N} \times \mathbb{R}^d \to \mathbb{R}^o$, named the \textbf{transition function} and the \textbf{output function}. In analogy to a graph convolution layer, the transition function defines a rule for updating vertex representations by aggregating transformations over representations of neighbor vertices. The vertex representation $\vec{\mathbf{x_i}}^{(t+1)}$ for vertex $i$ at time $(t+1)$ is computed as:
\begin{equation}\label{eq:GNN-model}
    \vec{\mathbf{x_i}}^{(t+1)} = \sum_{j \in \mathcal{N}(i)} h\left(l_i, l_j, l_{ij}, \vec{\mathbf{x_i}}^{(t)}\right)
\end{equation}
Where $l_i$, $l_j$ and $l_{ij}$ are respectively the labels for nodes $i$ and $j$ and edge $ij$ and $\mathbb{R}^d, \mathbb{R}^o$ are respectively the space of vertex representations and the output space. The model is defined over labelled graphs, but can still be implemented for unlabelled ones by supressing $l_i, l_j, l_{ij}$ from the transition function. After a certain number of iterations one should expect that vertex embeddings $\vec{\mathbf{x_i}}^{(t+1)}$ are enriched with structural information about the input graph. At this point, the output function $g$ can be used to compute an output for each vertex, given its final representation:
%\begin{equation}
$\vec{\mathbf{o_i}} = g(l_i, \vec{\mathbf{x_i}})$
%\end{equation}

In other words, the output at the end of the process is a set of $|\mathbf{V}|$ vectors $\in \mathbb{R}^o$. This is useful for node classification tasks, in which one can have $o$ equal the number of node classes and enforce $\vec{\mathbf{o_i}}$ to encode a probability distribution by incorporating a softmax layer into the output function $g$. If one would like to learn a function over the entire graph instead of its neighbors, there are many possibilities, of which one is to compute the output on an aggregation over all final vertex representations:
%\begin{equation}
    $\vec{\mathbf{o}} = g\left(\sum_{i \in \mathbf{V}}\vec{\mathbf{x_{v_g}}}\right)$
%\end{equation}
\subsection{Message-passing Neural Networks}
Message-passing neural networks implement a slight modification over the original GNN model, which is to define a specialized \textbf{update} function $u: \mathbb{R}^d \times \mathbb{R}^d \to \mathbb{R}^d$ to update the representation for vertex $i$ given its current representation and an aggregation $\mathbf{m_i}$ over transformations of neighbor vertex embeddings (which are referred to as ``messages'', hence message-passing neural networks), as an example:
\begin{center}
   $ \vec{\mathbf{x_i}}^{(t+1)} = u\bigl(\vec{\mathbf{x_i}}^{(t)}, \sum_{j \in \mathcal{N}(i)} h\bigl(l_i, l_j, l_{ij}, \vec{\mathbf{x_i}}^{(t)}\bigr)\bigr)$
\end{center}
Also, the update procedure is run over a fixed number of steps and it is usual to implement $u$ if using some type of recurrent network, such as Long-Short Term Memory (LSTM) cells \cite{LSTM,selsam2019neurosat}, or Gated Recurrent Units.
\subsection{Graph Attention Networks}
The Graph Attention Networks (GAT) \cite{velivckovic2017graph} augment models in the graph neural network family with an attention mechanism enabling vertices to weigh neighbor representations during their aggregation. As with other types of attention, a parameterized function is used to compute the weights dynamically, which enables the model to learn to weigh representations wisely.
The goal of the GAT is to compute a coefficient $e_{ij}: \mathbb{R}$ for each neighbor $j$ of a given vertex $i$, so that the aggregation in Equation \ref{eq:GNN-model} becomes:
\begin{center}
   $ \vec{\mathbf{x_i}}^{(t+1)} = \sum_{j \in \mathcal{N}(i)} e_{ij} h\bigl(l_i, l_j, l_{ij}, \vec{\mathbf{x_i}}^{(t)}\bigr)$\\
\end{center} 
To compute $e_{ij}$, the GAT introduces a weight matrix $\mathbf{W} \in \mathbb{R}^d \times \mathbb{R}^d$, used to multiply vertex embeddings for $i$ and $j$, which are concatenated and multiplied by a parameterized weight vector $\vec{\mathbf{a}}$. Finally, a non-linearity is applied to the computation in the above equation and then a softmax over the set of neighbors $\mathcal{N}(i)$ is applied over the exponential of the result, yielding:
%\begin{equation}
    $e_{ij} = \textrm{softmax}_j(\sigma(\vec{\mathbf{a}} \cdot (\mathbf{W}\vec{\mathbf{x_i}} || \mathbf{W}\vec{\mathbf{x_j}})))$ \\
%\end{equation}
GAT are known to outperform typical GCN architectures for graph classification tasks, as shown in the original paper \cite{velivckovic2017graph}.

%\subsection{Relationship to other Architectures}
%\subsubsection{Convolutional Neural Networks}
%\subsubsection{Recurrent Neural Networks}
%\subsubsection{Attention Mechanisms}

\section{Perspectives and Applications of GNNs to Neural-Symbolic Computing}
In this paper, we have seen that GNNs endowed with attention mechanisms are a promising direction of research towards the provision of rich reasoning and learning in type 6 neural-symbolic systems. Future work includes, of course, application and systematic evaluation of relevant specific tasks and data sets. These include what John McCarthy described as \emph{drosophila} tasks for Computer Science: basic problems that can illustrate the value of a computational model. 

Examples in the case of GNNs and NSC could be: 
(1) extrapolation of a learned classification of graphs as Hamiltonian to graphs of arbitrary size, (2) reasoning about a learned graph structure to generalise beyond the distribution of the training data, (3) reasoning about the $partOf(X,Y)$ relation to make sense of handwritten MNIST digits and non-digits. (4) using an adequate self-attention mechanism to make combinatorial reasoning computationally efficient. This last task relates to satisfiability including work on using GNNs to solve the TSP problem. The other tasks are related to meta-transfer learning across domains, extrapolation and causality. In terms of domains of application, the following are relevant.
\subsection{Relational Learning and Reasoning}
GNN models have been successfully applied to a number of relational reasoning tasks. Despite the success of convolutional networks, visual scene understanding is still out of reach for pure CNN models, and hence are a fertile ground for GNN-based models. Hybrid CNN + GNN models in particular have been successful in these tasks, having been applied to understanding human-object interactions, localising objects, and challenging visual question answering problems \cite{Santoro_2017}.
Relational reasoning has also been applied to physics, with models for extracting objects and relations in unsupervised fashion \cite{steenkiste2018relational}. GNNs coupled with differentiable ODE solvers have been used to learn the Hamiltonian dynamics of physical systems given their interactions modelled as a dynamic graph \cite{greydanus2019hamiltonian}.
The application of NSC models to life sciences is very promising, as graphs are natural representations for molecules, including proteins. In this context, \cite{STOKES2020688} have generated the first machine-learning discovered antibiotic (``halcin'') by training a GNN to predict the probability that a given input molecule has a growth inhibition effect on the bacterium E. coli and using it to rank randomly-generated molecules. Protein Structure Prediction, which is concerned with predicting the three-dimensional structure of proteins given its molecular description, is another promising problem for graph-based and NSC models such as DeepMind's AlphaFold and its variations \cite{wei2019protein}. 
%Some graph-based architectures have also been proposed for learning transformations over molecules, for example for the problem of molecular optimization, with some using an encoder-decoder architecture \cite{jin2018learning}.while others using reinforcement learning for goal-directed molecule generation \cite{you2018graph}.

In Natural language processing, tasks are usually defined over sequential data, but modelling textual data with graphs offers a number of advantages. Several approaches have defined graph neural networks over graphs of text co-occurrences, showing that these architectures improve upon the state-of-the-art for \emph{seq2seq} models \cite{yao2019graph}. GNN models have also been successfully applied to relational tasks over knowledge bases, such as link prediction \cite{schlichtkrull2018modeling}. As previously mentioned, attention mechanisms, which can be seen as a variation of models in the GNN family, have enabled substantial improvements in several NLP tasks through transfer learning over pretrained transformer language models \cite{devlin2018bert}. The extent to which language models pretrained over huge amounts of data can perform language understanding however is substantially debated, as pointed out by both Marcus \cite{marcus2020} and Kahneman \cite{fireside2020}.

Graph-based neural network models have also found a fertile field of application in software engineering: due to the structured and unambiguous nature of code, it can be represented naturally with graphs that are derived unambiguously via parsing. Several works have then utilised GNNs to perform analysis over graph representations of programs and obtained significant results. 
More specifically, Microsoft's ``Deep Program Understanding" research program has used a GNN variant called Gated Graph Sequence Neural Networks \cite{li2016gated} in a large number of applications, including spotting errors, suggesting variable names, code completion \cite{grockschmidt2019gencode}, as well as edit representation and automatically applying edits to programs \cite{yin2019edits}.

\subsection{Combinatorial Optimization and Constraint Satisfaction Problems}
Many combinatorial optimization problems are relational in structure and thus are prime application targets to GNN-based models \cite{bengio2018tourdhorizon}. For instance,  \cite{khalil2017combinatorial} uses a GNN-like model to embed graphs and use these embeddings in their heuristic search for the Minimum Vertex Cover (MVC), Maximum Cut and Traveling Salesperson (TSP) problems.
Regarding end-to-end models, \cite{kool2019attention} trained a transformer-based GNN model to embed TSP answers and extract solutions with an attention-based decoder, while obtaining better performance than previous work. \cite{li2018combinatorial} used a GCN as a heuristic to a search algorithm, applying this method on four canonical NP-complete problems, namely Maximal Independent Set, MVC, Maximal Clique, and the Boolean Satisfiability Problem (SAT).  \cite{palm2018recurrentrelational} achieved convergent algorithms over relational problems. The expressiveness of GNNs has also been the focus of recent research \cite{sato2020}.

Regarding $\mathcal{NP}$-Hard problems, neural-symbolic models with an underlying GNN formalization have been proposed to train solvers for the decision variants of the SAT, TSP and graph colouring problems, respectively \cite{selsam2019neurosat,prates2019tsp,lemos2019gcp}. This allowed these models to be trained with a single bit of supervision on each instance, \cite{selsam2019neurosat,leytonAAAI2020} being able to extract assignments from the trained model, \cite{prates2019tsp} performing a binary search on the prediction probability to estimate the optimal route cost. \cite{toenshoff2019runcsp} built an end-to-end framework for dealing with (boolean) constraint satisfaction problems in general, extending the previous works and providing comparisons and performance increases, and \cite{abboud2020} have proposed a GNN-based architecture that learns to approximate DNF counting. There has also been work in generative models for combinatorial optimization, such as \cite{you2019g2sat}, which generates SAT instances using a graph-based approach.
%There has also been work in generative models for combinatorial optimization problems, such as \cite{you2019g2sat}, which generates SAT instances using a graph-based approach.
\section{Conclusions}
We presented a review on the relationship between Graph Neural Network (GNN) models and  architectures and Neural-Symbolic Computing (NSC). In order to do so, we presented the main recent research results that highlight the potential applications of these related fields both in foundational and applied AI and Computer Science problems.
The interplay between the two fields is beneficial to several areas. These range from combinatorial optimization/constraint satisfaction to relational reasoning, which has been the subject of increasing industrial relevance in natural language processing, life sciences and computer vision and image understanding \cite{raghavan19,marcus2020}. This is largely due to the fact that many learning tasks can be easily and naturally captured using graph representations, which can be seen as a generalization over the traditional sequential (RNN) and grid-based representations (CNN) in the family of deep learning building blocks. Finally, it is worth mentioning that the principled integration of both methodologies (GNN and NSC) offers a richer alternative to the construction of trustful, explainable and robust AI systems, which is clearly an invaluable research endeavor. 
\section*{Acknowledgements}
This work is partly supported by CNPq and CAPES, Brazil - Finance Code 001. Moshe Vardi is supported in part by NSF grants IIS-1527668, CCF-1704883,
IIS-1830549, DoD MURI grant N00014-20-1-2787,
and an award from the Maryland Procurement Office. 
%\subsubsection{Reinforcement Learning}

%RL for comb opt \cite{khalil2017combinatorial}

%RL for robot control \cite{sanchez-gonzalez2018graph,wang2018nervenet}

%\subsubsection{Semi-supervised, Few-shot and Zero-shot Learning}

% No need for clearpage, IJCAI 2019 had the same rules for paper length
% https://www.ijcai19.org/call-for-survey.html
% And here is an accepted survey which had references starting on the same page the text ends:
% https://www.ijcai.org/Proceedings/2019/0874.pdf
% https://www.ijcai.org/Proceedings/2019/0876.pdf
\bibliographystyle{named}
\bibliography{ijcai20.bib}

\begin{thebibliography}{}

\bibitem[\protect\citeauthoryear{Abboud \bgroup \em et al.\egroup
  }{2020}]{abboud2020}
Ralph Abboud, Ismail Ceylan, and Thomas Lukasiewicz.
\newblock Learning to reason: Leveraging neural networks for approximate {DNF}
  counting.
\newblock In {\em AAAI}, pages 1--6, 2020.

\bibitem[\protect\citeauthoryear{Arabshahi \bgroup \em et al.\egroup
  }{2018}]{ArabshahiSA18}
Forough Arabshahi, Sameer Singh, and Animashree Anandkumar.
\newblock Combining symbolic expressions and black-box function evaluations in
  neural programs.
\newblock In {\em {ICLR}}, 2018.

\bibitem[\protect\citeauthoryear{Arabshahi \bgroup \em et al.\egroup
  }{2019}]{Arabshahi19}
Forough Arabshahi, Zhichu Lu, Sameer Singh, and Animashree Anandkumar.
\newblock Memory augmented recursive neural networks.
\newblock {\em CoRR}, abs/1911.01545, 2019.

\bibitem[\protect\citeauthoryear{Battaglia \bgroup \em et al.\egroup
  }{2018}]{battaglia2018relational}
Peter Battaglia, Jessica Hamrick, Victor Bapst, Alvaro Sanchez{-}Gonzalez,
  Vin{\'{\i}}cius~Zambaldi et. al.~%and %Mateusz~Malinowski, %Andrea Tacchetti,
  %David Raposo, %Adam Santoro, %Ryan Faulkner,
  %~{\c{C}}aglar~G{\"{u}}l{\c{c}}ehre %and %H. Francis~Song, %Andrew Ballard,
  %Justin Gilmer, %George Dahl, %Ashish Vaswani, %Kelsey Allen, %Charles Nash,
  %Victoria Langston, %~Chris Dyer, %~Nicolas Heess, %~Daan Wierstra,
  %~Pushmeet Kohli, %~Matthew Botvinick, %~Oriol Vinyals, %~Yujia Li, and
  %~Razvan Pascanu.
\newblock Relational inductive biases, deep learning, and graph networks.
\newblock {\em CoRR}, abs/1806.01261, 2018.

\bibitem[\protect\citeauthoryear{Bengio \bgroup \em et al.\egroup
  }{2018}]{bengio2018tourdhorizon}
Yoshua Bengio, Andrea Lodi, and Antoine Prouvost.
\newblock Machine learning for combinatorial optimization: a methodological
  tour d'horizon.
\newblock {\em CoRR abs/1811.06128}, 2018.

\bibitem[\protect\citeauthoryear{Bordes \bgroup \em et al.\egroup
  }{2011}]{Bordes_2011}
Antoine Bordes, Jason Weston, Ronan Collobert, and Yoshua Bengio.
\newblock Learning structured embeddings of knowledge bases.
\newblock In {\em AAAI}, 2011.

\bibitem[\protect\citeauthoryear{Brockschmidt \bgroup \em et al.\egroup
  }{2019}]{grockschmidt2019gencode}
Marc Brockschmidt, Miltiadis Allamanis, Alexander Gaunt, and Oleksandr Polozov.
\newblock Generative code modeling with graphs.
\newblock In {\em {ICLR}}, 2019.

\bibitem[\protect\citeauthoryear{Cameron \bgroup \em et al.\egroup
  }{2020}]{leytonAAAI2020}
Chris Cameron, Rex Chen, Jason Hartford, and Kevin Leyton-Brown.
\newblock Predicting propositional satisfiability via end-to-end learning.
\newblock In {\em AAAI}, 2020.

\bibitem[\protect\citeauthoryear{Chami \bgroup \em et al.\egroup
  }{2020}]{chami2020}
Ines Chami, Sami Abu{-}El{-}Haija, Bryan Perozzi, Christopher R{\'{e}}, and
  Kevin Murphy.
\newblock Machine learning on graphs: {A} model and comprehensive taxonomy.
\newblock {\em CoRR}, abs/2005.03675, 2020.

\bibitem[\protect\citeauthoryear{d'Avila Garcez and Lamb}{2003}]{NIPS03}
Artur d'Avila Garcez and Luís Lamb.
\newblock Reasoning about time and knowledge in neural symbolic learning
  systems.
\newblock In {\em {NIPS}}, pages 921--928, 2003.

\bibitem[\protect\citeauthoryear{{d{'}Avila Garcez} \bgroup \em et al.\egroup
  }{2009}]{garcez_book2}
{Artur} {d{'}Avila Garcez}, Luís~C. Lamb, and Dov~M. Gabbay.
\newblock {\em Neural-Symbolic Cognitive Reasoning}.
\newblock Springer, 2009.

\bibitem[\protect\citeauthoryear{d'Avila Garcez \bgroup \em et al.\egroup
  }{2015}]{garcez2015}
Artur d'Avila Garcez, Tarek Besold, Luc de~Raedt, Peter F{\"{o}}ldi{\'{a}}k,
  Pascal Hitzler, Thomas Icard, Kai-Uwe K{\"{u}}hnberger, Luís~C. Lamb, Risto
  Miikkulainen, and Daniel Silver.
\newblock Neural-symbolic learning and reasoning: Contributions and challenges.
\newblock In {\em {AAAI} Spring Symposia}, 2015.

\bibitem[\protect\citeauthoryear{d'Avila Garcez \bgroup \em et al.\egroup
  }{2019}]{JAL19}
Artur d'Avila Garcez, Marco Gori, Luís~C. Lamb, Luciano Serafini, Michael
  Spranger, and Son Tran.
\newblock Neural-symbolic computing: An effective methodology for principled
  integration of machine learning and reasoning.
\newblock {\em {FLAP}}, 6(4):611--632, 2019.

\bibitem[\protect\citeauthoryear{Davis}{2019}]{ernie}
Ernest Davis.
\newblock The use of deep learning for symbolic integration: {A} review of
  ({Lample and Charton}, 2019).
\newblock {\em CoRR}, abs/1912.05752, 2019.

\bibitem[\protect\citeauthoryear{Devlin \bgroup \em et al.\egroup
  }{2019}]{devlin2018bert}
Jacob Devlin, Ming-Wei Chang, Kenton Lee, and Kristina Toutanova.
\newblock {BERT}: Pre-training of deep bidirectional transformers for language
  understanding.
\newblock In {\em {NAACL-HLT}}, pages 4171--4186, 2019.

\bibitem[\protect\citeauthoryear{Diligenti \bgroup \em et al.\egroup
  }{2017}]{DiligentiGS17}
Michelangelo Diligenti, Marco Gori, and Claudio Saccà.
\newblock Semantic-based regularization for learning and inference.
\newblock {\em Artif. Intell.}, 244:143--165, 2017.

\bibitem[\protect\citeauthoryear{Evans and Grefenstette}{2018}]{Evans_18}
Richard Evans and Edward Grefenstette.
\newblock Learning explanatory rules from noisy data.
\newblock {\em JAIR}, 61:1--64, 2018.

\bibitem[\protect\citeauthoryear{Galassi \bgroup \em et al.\egroup
  }{2020}]{kersting2019}
Andrea Galassi, Kristian Kersting, Marco Lippi, Xiaoting Shao, and Paolo
  Torroni.
\newblock Neural-symbolic argumentation mining: An argument in favor of deep
  learning and reasoning.
\newblock {\em Front. Big Data}, 2:52, 2020.

\bibitem[\protect\citeauthoryear{Garcia and Bruna}{2018}]{garcia18}
Victor Garcia and Joan Bruna.
\newblock Few-shot learning with graph neural networks.
\newblock In {\em {ICLR}}, pages 1--13, 2018.

\bibitem[\protect\citeauthoryear{Goyal \bgroup \em et al.\egroup
  }{2019}]{goyal2019}
Anirudh Goyal, Alex Lamb, Jordan Hoffmann, Shagun Sodhani, Sergey Levine,
  Yoshua Bengio, and Bernhard Sch{\"{o}}lkopf.
\newblock Recurrent independent mechanisms.
\newblock {\em CoRR}, abs/1909.10893, 2019.

\bibitem[\protect\citeauthoryear{Greydanus \bgroup \em et al.\egroup
  }{2019}]{greydanus2019hamiltonian}
Samuel Greydanus, Misko Dzamba, and Jason Yosinski.
\newblock Hamiltonian neural networks.
\newblock In {\em NeurIPS}, pages 5353--15363, 2019.

\bibitem[\protect\citeauthoryear{Hitzler \bgroup \em et al.\egroup
  }{2004}]{Hitzler04}
Pascal Hitzler, Steffen H{\"{o}}lldobler, and Anthony~K. Seda.
\newblock Logic programs and connectionist networks.
\newblock {\em J. Appl. Log.}, 2(3):245--272, 2004.

\bibitem[\protect\citeauthoryear{Hochreiter and Schmidhuber}{1997}]{LSTM}
Sepp Hochreiter and J{\"{u}}rgen Schmidhuber.
\newblock Long short-term memory.
\newblock {\em Neural Computation}, 9(8):1735--1780, 1997.

\bibitem[\protect\citeauthoryear{Huang \bgroup \em et al.\egroup
  }{2017}]{Smolensky}
Qiuyuan Huang, Paul Smolensky, Xiaodong He, Li~Deng, and Dapeng~Oliver Wu.
\newblock A neural-symbolic approach to natural language tasks.
\newblock {\em CoRR}, abs/1710.11475, 2017.

\bibitem[\protect\citeauthoryear{Kahneman \bgroup \em et al.\egroup
  }{2020}]{fireside2020}
Daniel Kahneman, Francesca Rossi, Geoffrey Hinton, Yoshua Bengio, and Yann
  LeCun.
\newblock {AAAI20} fireside chat with {D}aniel {K}ahneman.
\newblock \url{https://vimeo.com/390814190?ref=tw-share}, 2020.
\newblock Accessed 23/02/2020.

\bibitem[\protect\citeauthoryear{Khalil \bgroup \em et al.\egroup
  }{2017}]{khalil2017combinatorial}
Elias Khalil, Hanjun Dai, Yuyu Zhang, Bistra Dilkina, and Le~Song.
\newblock Learning combinatorial optimization algorithms over graphs.
\newblock In {\em {NIPS}}, pages 6348--6358, 2017.

\bibitem[\protect\citeauthoryear{Kipf and Welling}{2017}]{kipf2016semi}
Thomas~N. Kipf and Max Welling.
\newblock Semi-supervised classification with graph convolutional networks.
\newblock In {\em ICLR}, 2017.

\bibitem[\protect\citeauthoryear{Kool \bgroup \em et al.\egroup
  }{2019}]{kool2019attention}
Wouter Kool, Herke van Hoof, and Max Welling.
\newblock Attention, learn to solve routing problems!
\newblock In {\em {ICLR}}, 2019.

\bibitem[\protect\citeauthoryear{Lample and Charton}{2020}]{Lample2020Deep}
Guillaume Lample and François Charton.
\newblock Deep learning for symbolic mathematics.
\newblock In {\em {ICLR}}, 2020.

\bibitem[\protect\citeauthoryear{LeCun \bgroup \em et al.\egroup
  }{2015}]{Hinton-nature}
Yann LeCun, Yoshua Bengio, and Geoffrey Hinton.
\newblock Deep learning.
\newblock {\em Nature}, 521(7553):436--444, 2015.

\bibitem[\protect\citeauthoryear{Lemos \bgroup \em et al.\egroup
  }{2019}]{lemos2019gcp}
Henrique Lemos, Marcelo Prates, Pedro Avelar, and Luís~C. Lamb.
\newblock Graph colouring meets deep learning: Effective graph neural network
  models for combinatorial problems.
\newblock In {\em {ICTAI}}, pages 879--885, 2019.

\bibitem[\protect\citeauthoryear{Li \bgroup \em et al.\egroup
  }{2016}]{li2016gated}
Yujia Li, Daniel Tarlow, Marc Brockschmidt, and Richard Zemel.
\newblock Gated graph sequence neural networks.
\newblock In {\em {ICLR}}, 2016.

\bibitem[\protect\citeauthoryear{Li \bgroup \em et al.\egroup
  }{2018}]{li2018combinatorial}
Zhuwen Li, Qifeng Chen, and Vladlen Koltun.
\newblock Combinatorial optimization with graph convolutional networks and
  guided tree search.
\newblock In {\em NeurIPS}, 2018.

\bibitem[\protect\citeauthoryear{Manhaeve \bgroup \em et al.\egroup
  }{2018}]{Robin_2018}
Robin Manhaeve, Sebastijan Dumancic, Angelika Kimmig, Thomas Demeester, and
  Luc~De Raedt.
\newblock Deepproblog: Neural probabilistic logic programming.
\newblock In {\em NeurIPS}, 2018.

\bibitem[\protect\citeauthoryear{Mao \bgroup \em et al.\egroup
  }{2019}]{Mao_2019}
Jiayuan Mao, Chuang Gan, Pushmeet Kohli, Joshua Tenenbaum, and Jiajun Wu.
\newblock {The Neuro-Symbolic Concept Learner: Interpreting scenes, words, and
  sentences from natural supervision}.
\newblock In {\em ICLR}, 2019.

\bibitem[\protect\citeauthoryear{Marcus}{2020}]{marcus2020}
Gary Marcus.
\newblock The next decade in {AI}: {F}our steps towards robust artificial
  intelligence.
\newblock {\em CoRR}, abs/1801.00631, 2020.

\bibitem[\protect\citeauthoryear{Palm \bgroup \em et al.\egroup
  }{2018}]{palm2018recurrentrelational}
Rasmus Palm, Ulrich Paquet, and Ole Winther.
\newblock Recurrent relational networks.
\newblock In {\em NeurIPS}, pages 3372--3382, 2018.

\bibitem[\protect\citeauthoryear{Prates \bgroup \em et al.\egroup
  }{2019}]{prates2019tsp}
Marcelo Prates, Pedro Avelar, Henrique Lemos, Luís Lamb, and Moshe Vardi.
\newblock Learning to {S}olve {NP}-{C}omplete {P}roblems: {A} {G}raph {N}eural
  {N}etwork for {D}ecision {TSP}.
\newblock In {\em {AAAI-2019}}, pages 4731--4738, 2019.

\bibitem[\protect\citeauthoryear{Raghavan}{2019}]{raghavan19}
Sriram Raghavan.
\newblock 2020 {AI} predictions from {IBM} research.
\newblock \url{https://www.ibm.com/blogs/research/2019/12/2020-ai-predictions},
  2019.
\newblock Accessed 20/02/2020.

\bibitem[\protect\citeauthoryear{Rockt{\"{a}}schel and
  Riedel}{2016}]{Rocktaschel_2016}
Tim Rockt{\"{a}}schel and Sebastian Riedel.
\newblock Learning knowledge base inference with neural theorem provers.
\newblock In {\em {AKBC@NAACL-HLT}}, 2016.

\bibitem[\protect\citeauthoryear{Santoro \bgroup \em et al.\egroup
  }{2017}]{Santoro_2017}
Adam Santoro, David Raposo, David Barrett, Mateusz Malinowski, Razvan Pascanu,
  Peter Battaglia, and Tim Lillicrap.
\newblock A simple neural network module for relational reasoning.
\newblock In {\em NIPS}, pages 4967--4976, 2017.

\bibitem[\protect\citeauthoryear{Sato}{2020}]{sato2020}
Ryoma Sato.
\newblock A survey on the expressive power of graph neural networks.
\newblock {\em CoRR}, abs/2003.04078, 2020.

\bibitem[\protect\citeauthoryear{Scarselli \bgroup \em et al.\egroup
  }{2008}]{scarselli2008graph}
Franco Scarselli, Marco Gori, Ah~Tsoi, Markus Hagenbuchner, and Gabriele
  Monfardini.
\newblock The graph neural network model.
\newblock {\em IEEE T Neural Networ}, 20(1):61--80, 2008.

\bibitem[\protect\citeauthoryear{Schlichtkrull \bgroup \em et al.\egroup
  }{2018}]{schlichtkrull2018modeling}
Michael Schlichtkrull, Thomas Kipf, Peter Bloem, Rianne van~den Berg, Ivan
  Titov, and Max Welling.
\newblock Modeling relational data with graph convolutional networks.
\newblock {\em {ESWC}}, pages 593--607, 2018.

\bibitem[\protect\citeauthoryear{Selsam \bgroup \em et al.\egroup
  }{2019}]{selsam2019neurosat}
Daniel Selsam, Matthew Lamm, Benedikt B{\"{u}}nz, Percy Liang, Leonardo
  de~Moura, and David~L. Dill.
\newblock Learning a {SAT} solver from single-bit supervision.
\newblock In {\em {ICLR}}, pages 1--11, 2019.

\bibitem[\protect\citeauthoryear{Serafini and d'Avila Garcez}{2016}]{LTN}
Luciano Serafini and Artur d'Avila Garcez.
\newblock Logic tensor networks: Deep learning and logical reasoning from data
  and knowledge.
\newblock {\em CoRR}, abs/1606.04422, 2016.

\bibitem[\protect\citeauthoryear{Stokes \bgroup \em et al.\egroup
  }{2020}]{STOKES2020688}
Jonathan~M. Stokes, Kevin Yang, Kyle Swanson, and Wengong~Jin et~al.
\newblock A deep learning approach to antibiotic discovery.
\newblock {\em Cell}, 180, 2020.

\bibitem[\protect\citeauthoryear{Sutskever and Hinton}{2008}]{Ilya_2008}
Ilya Sutskever and Geoffrey Hinton.
\newblock Using matrices to model symbolic relationship.
\newblock In {\em NIPS}, pages 1593--1600, 2008.

\bibitem[\protect\citeauthoryear{Toenshoff \bgroup \em et al.\egroup
  }{2019}]{toenshoff2019runcsp}
Jan Toenshoff, Martin Ritzert, Hinrikus Wolf, and Martin Grohe.
\newblock {RUN-CSP:} unsupervised learning of message passing networks for
  binary constraint satisfaction problems.
\newblock {\em CoRR}, abs/1909.08387, 2019.

\bibitem[\protect\citeauthoryear{Townsend \bgroup \em et al.\egroup
  }{2019}]{Joe19}
Joe Townsend, Thomas Chaton, and João Monteiro.
\newblock Extracting relational explanations from deep neural networks: A
  survey from a neural-symbolic perspective.
\newblock {\em IEEE T Neur Net Learn}, pages 1--15, 2019.

\bibitem[\protect\citeauthoryear{van Harmelen and ten
  Teije}{2019}]{van2019boxology}
Frank van Harmelen and Annette ten Teije.
\newblock A boxology of design patterns for hybrid learning and reasoning
  systems.
\newblock {\em J Web Eng}, 18(1):97--124, 2019.

\bibitem[\protect\citeauthoryear{van Steenkiste \bgroup \em et al.\egroup
  }{2018}]{steenkiste2018relational}
Sjoerd van Steenkiste, Michael Chang, Klaus Greff, and J{\"{u}}rgen
  Schmidhuber.
\newblock Relational neural expectation maximization: Unsupervised discovery of
  objects and their interactions.
\newblock In {\em {ICLR}}, 2018.

\bibitem[\protect\citeauthoryear{Veli\v{c}kovi\'c \bgroup \em et al.\egroup
  }{2018}]{velivckovic2017graph}
Petar Veli\v{c}kovi\'c, Guillem Cucurull, Arantxa Casanova, Adriana Romero,
  Pietro Li{\`{o}}, and Yoshua Bengio.
\newblock Graph attention networks.
\newblock In {\em {ICLR}}, 2018.

\bibitem[\protect\citeauthoryear{Vinyals \bgroup \em et al.\egroup
  }{2015}]{vinyals2015pointer}
Oriol Vinyals, Meire Fortunato, and Navdeep Jaitly.
\newblock Pointer networks.
\newblock In {\em {NIPS}}, pages 2692--2700, 2015.

\bibitem[\protect\citeauthoryear{Wei}{2019}]{wei2019protein}
Guo-Wei Wei.
\newblock Protein structure prediction beyond alphafold.
\newblock {\em Nature Mach. Intell.}, 1:336--337, 2019.

\bibitem[\protect\citeauthoryear{Wu \bgroup \em et al.\egroup
  }{2019}]{wu2019comprehensive}
Zonghan Wu, Shirui Pan, Fengwen Chen, Guodong Long, Chengqi Zhang, and
  Philip~S. Yu.
\newblock A comprehensive survey on graph neural networks.
\newblock {\em CoRR}, abs/1901.00596, 2019.

\bibitem[\protect\citeauthoryear{Yao \bgroup \em et al.\egroup
  }{2019}]{yao2019graph}
Liang Yao, Chengsheng Mao, and Yuan Luo.
\newblock Graph convolutional networks for text classification.
\newblock In {\em AAAI}, pages 7370--7377, 2019.

\bibitem[\protect\citeauthoryear{Yin \bgroup \em et al.\egroup
  }{2019}]{yin2019edits}
Pengcheng Yin, Graham Neubig, Miltiadis Allamanis, and Alexander~Gaunt
  Marc~Brockschmidt.
\newblock Learning to represent edits.
\newblock In {\em {ICLR}}, 2019.

\bibitem[\protect\citeauthoryear{You \bgroup \em et al.\egroup
  }{2019}]{you2019g2sat}
Jiaxuan You, Haoze Wu, Clark Barrett, Raghuram Ramanujan, and Jure Leskovec.
\newblock {G2SAT:} learning to generate {SAT} formulas.
\newblock In {\em NeurIPS}, pages 10552--10563, 2019.

\bibitem[\protect\citeauthoryear{Zhang \bgroup \em et al.\egroup
  }{2018}]{Zhang2018}
Ziwei Zhang, Peng Cui, and Wenwu Zhu.
\newblock Deep learning on graphs: {A} survey.
\newblock {\em CoRR}, abs/1812.04202, 2018.

\end{thebibliography}
\end{document}